\documentclass{article}
\usepackage[utf8]{inputenc}
\usepackage{amsmath}
\usepackage{hyperref}
\usepackage{graphicx}
\usepackage{subcaption}
\usepackage{amssymb}
\usepackage{listings}
\usepackage{amsthm}
\usepackage{algorithm}
\usepackage[noend]{algpseudocode}
\usepackage{comment} 
\usepackage{amsmath,amssymb}
\usepackage{amsthm}
\usepackage{changepage}
\usepackage{cite}
\usepackage{nameref}
\usepackage{graphicx}
\usepackage{subcaption}
\usepackage{array}
\usepackage{authblk}
\usepackage{multirow}
\usepackage{xcolor}
\usepackage{comment}
\usepackage{color}

\usepackage{amsthm}
\usepackage{amssymb}

\theoremstyle{definition}

\theoremstyle{remark}

\usepackage[margin=0.9in]{geometry}

\usepackage{comment} 

\title{Knowledge-integrated AutoEncoder Model}
\author[1]{Teddy Lazebnik\footnote{Corresponding author: t.lazebnik@ucl.ac.uk}}
\author[2]{Liron Simon-Keren}
\affil[1]{Department of Cancer Biology, Cancer Institute, University College London, London, UK}
\affil[2]{School of Mechanical Engineering, Tel Aviv University, Israel}

\date{}

\def \methodName {KiAE}
\def \methodNamefull {Knowledge-integrated AutoEncoder}

\begin{document}
 
\maketitle

\begin{abstract}
Data encoding is a common and central operation in most data analysis tasks. The performance of other models, downstream in the computational process, highly depends on the quality of data encoding.
One of the most powerful ways to encode data is using the neural network AutoEncoder (AE) architecture. However, the developers of AE are not able to easily influence the produced embedding space, as it is usually treated as a \textit{black box} technique, which makes it uncontrollable and not necessarily has desired properties for downstream tasks. In this paper, we introduce a novel approach for developing AE models that can integrate external knowledge sources into the learning process, possibly leading to more accurate results. The proposed \methodNamefull{} (\methodName{}) model is able to leverage domain-specific information to make sure the desired distance and neighborhood properties between samples are preservative in the embedding space. The proposed model is evaluated on three large-scale datasets from three different scientific fields and is compared to nine existing encoding models. The results demonstrate that the \methodName{} model effectively captures the underlying structures and relationships between the input data and external knowledge, meaning it generates a more useful representation. This leads to outperforming the rest of the models in terms of reconstruction accuracy. \\

\noindent
\textbf{Keywords}: data-driven encoding; biologically-inspired loss function; expert-driven model.

\end{abstract}

\section{Introduction}
\label{sec:intro}
Data encoding is a crucial step in many data-driven analysis models across various fields, including economics, physics, and biology \cite{regression_1,regression_2,regression_3,regression_4,regression_5,regression_6,intro_phy_1,intro_phy_2,intro_eco_1,intro_eco_2}. Intuitively, the encoding process refers to the process of converting raw data into a standardized format that can be easily analyzed and interpreted \cite{data_encoding}. This process typically involves transforming data into a numerical space, usually with a dimension much smaller than the original one \cite{ls_1,ls_2,ls_3}. The encoded representation is used to create models or perform statistical analyses. As such, data encoding plays a critical role in several applications, including machine learning, image and speech recognition, natural language processing, and genomic analysis, among others \cite{ga_dna_1,similarity_general_2,speech,nlp}.

Recently, data-driven models are becoming increasingly complex, and as a result, the data used to train them is becoming more complex and large. As a result, there is a growing interest in the encoding space of the data and its properties \cite{latent_1,latent_2,latent_3}. In particular, \textit{latent spaces}, in the context of deep learning, refer to the encoded representations of data samples that are learned by neural networks (NN) during training. The term "latent" is used because these representations are not directly observable, but rather inferred from the training data by the NN \cite{latent_define}. Latent spaces are an essential component of several deep learning models, including AutoEncoders (AEs), generative adversarial networks, and transformers.

While providing promising results, current encoding methods leave the user (partially) blind to the properties of the encoded space that the model learned. This fact might result in unwanted behavior when known properties of the data or the domain are distributed. When focusing on the downstream tasks, users might be fine with such conditions when the results are promising but for debugging, optimization, and even explainability of the results, the current situation is sub-optimal.

In recent years, incorporating domain knowledge into machine learning models has gained significant momentum, as it has proven to be beneficial in many different applications. Incorporating domain knowledge, such as subject matter expertise or prior knowledge about the data or task, can lead to more accurate and interpretable results, as well as more efficient model designs. For example, in symbolic regression tasks, incorporating knowledge of physical laws or mathematical relationships, can reduce the computational resources required by minimizing the search space to only functions that fulfill a desired quality. Additionally, the added restraints over the search space aid in the discovery of a logical structure underlying the data \cite{scimed,sr_1,sr_2}. Another area where domain knowledge is being used is in automated machine learning (AutoML). Similar to the case of symbolic regression, domain knowledge can be used to guide the design of AutoML algorithms and to constrain the search space for the best model \cite{auto_sklearn}. Finally, domain knowledge can also be useful in the design of machine learning and deep learning models themselves \cite{ml_k_1,ml_k_2,ml_k_3}. By incorporating prior knowledge such as the structure of the input data or the relationships between different variables, models can be designed to be more efficient and accurate. This can be particularly useful in situations where data is limited, noisy, or difficult to collect.

In this paper, we present a novel approach for integrating domain knowledge into the \textit{Latent space} of AEs named \methodNamefull{} (\methodName{}). With this approach, we propose to use an AutoEncoder-based architecture that preserves domain knowledge in the form of distance and neighborhood properties between labeled groups in the dataset, even if these properties are only partially known from outside of the dataset, as a piece of domain knowledge.

We demonstrate that \methodName{} outperforms nine other AE models in three clustering tasks of datasets from the fields of economics, physics, and biology. Additionally, we demonstrate the drawback of the method, where, as in all other knowledge-informed models, the performance of the model is highly susceptible to the integration of incorrect knowledge. 

The rest of this paper is organized as follows: Section \ref{sec:related_work} reviews the state-of-the-art AE models with a focus on knowledge-integrated approaches. Section \ref{sec:methods} formally introduces \methodName{} and outlines the experimental setup used to evaluate it. Section \ref{sec:results} sets forth the experiments' results. Lastly, section \ref{sec:discussion}, summarizes our conclusions and discusses opportunities for future work.

\section{Related Work}
\label{sec:related_work}
Data encoding is the process of transforming raw data into a structured format that can be easily processed by a computer system \cite{data_encoding}. There are two main approaches to data encoding: rule-based and data-driven. Traditional rule-based encoding methods involve using a fixed set of rules to transform data into a specific format \cite{manual_encoding}. For instance, geographical locations are represented using manually defined latitude and longitude values \cite{geo_info}. In contrast, data-driven encoding techniques employ statistical models to learn the encoding scheme from the data itself. This approach is particularly useful in scenarios where traditional encoding methods are not feasible or when the data has complex structures and patterns that are difficult to capture with fixed rules \cite{driven_encoding}. However, rule-based encoding has the advantage of being more interpretable compared to data-driven encoding \cite{rule_data_encoding}. Data-driven encoding techniques can also optimize data representation for specific tasks by learning encoding schemes that capture the most important features of the data \cite{en_g_1,en_g_2}.

In recent years, AEs have gained popularity as powerful computational tools to achieve various goals. Specifically, AEs are a type of neural network that learn to encode and decode data in an unsupervised manner \cite{ae_review}. They consist of an encoder network that compresses the input data into a low-dimensional representation, also known as the \textit{latent space}, and a decoder network that reconstructs the original data from the compressed representation. AEs have found widespread use in various tasks such as image and audio processing, anomaly detection, and data compression. AEs are particularly useful for tasks where labeled data is scarce or expensive to obtain \cite{ae_review}. 

Due to their usefulness, AEs have found widespread use in various fields such as economics, physics, and biology. For example, in the economic domain, \cite{rw_economy_1} proposed a two-step electricity theft detection strategy that uses a convolutional autoencoder for electricity theft identification, where abnormal electricity consumption patterns are identified against the uniformity and periodicity of normal power consumption users. \cite{rw_economy_2} proposed an anomaly detection model for smart farming using an AE model that reconstructs normal data with a low reconstruction loss and anomalous data with a high loss. In the physics domain, AEs have also gained much popularity. For instance, \cite{rw_phy_1} investigated the usage of physics-constrained data-driven computing for material design using AEs. In addition, \cite{rw_phy_2} studied the tagging of top jet images in a background of QCD jet images using AE architectures, with similar results obtained by \cite{rw_phy_3}. Similarly, in the biological domain, \cite{rw_bio_1} reviewed several variational AEs in the context of gene expression, showing they outperform rule-based encoding methods even with a small amount of data. \cite{rw_bio_2} developed an algorithm that aids in the curation of gene annotations by automatically suggesting inaccuracies and predicting previously-unidentified gene functions, accelerating the rate of gene function discovery, which is based on AEs. The authors tested their AE model on gene annotation data from the Gene Ontology project, showing it outperforms many machine learning models.

The integration of domain knowledge into AEs is gaining popularity as an approach to enrich the dataset or direct the learning process \cite{ae_info_intro}. For example, \cite{ae_info_1} proposed a Multi-view Factorization AutoEncoder (MAE) with network constraints that can seamlessly integrate multi-omics data and domain knowledge such as molecular interaction networks. This method learns feature and patient embedding simultaneously, using deep representation learning that constrains both feature representations and patient representations to specific regularization terms in the training objective. \cite{ae_info_2} proposed a method to incorporate domain knowledge explicitly in the generation process to achieve the Semantically Adversarial Generation (SAG), focusing on the driving scenes encoding task. They first categorize domain knowledge into two types; the property of objects and the relationship among objects. This approach is implemented with a tree-structured variational AutoEncoder (T-VAE) to learn hierarchical scene representation.

\section{Methods and Materials}

\subsection{Model definition}
\label{sec:methods}
\begin{figure}[!ht]
    \centering
    \includegraphics[width=0.99\textwidth]{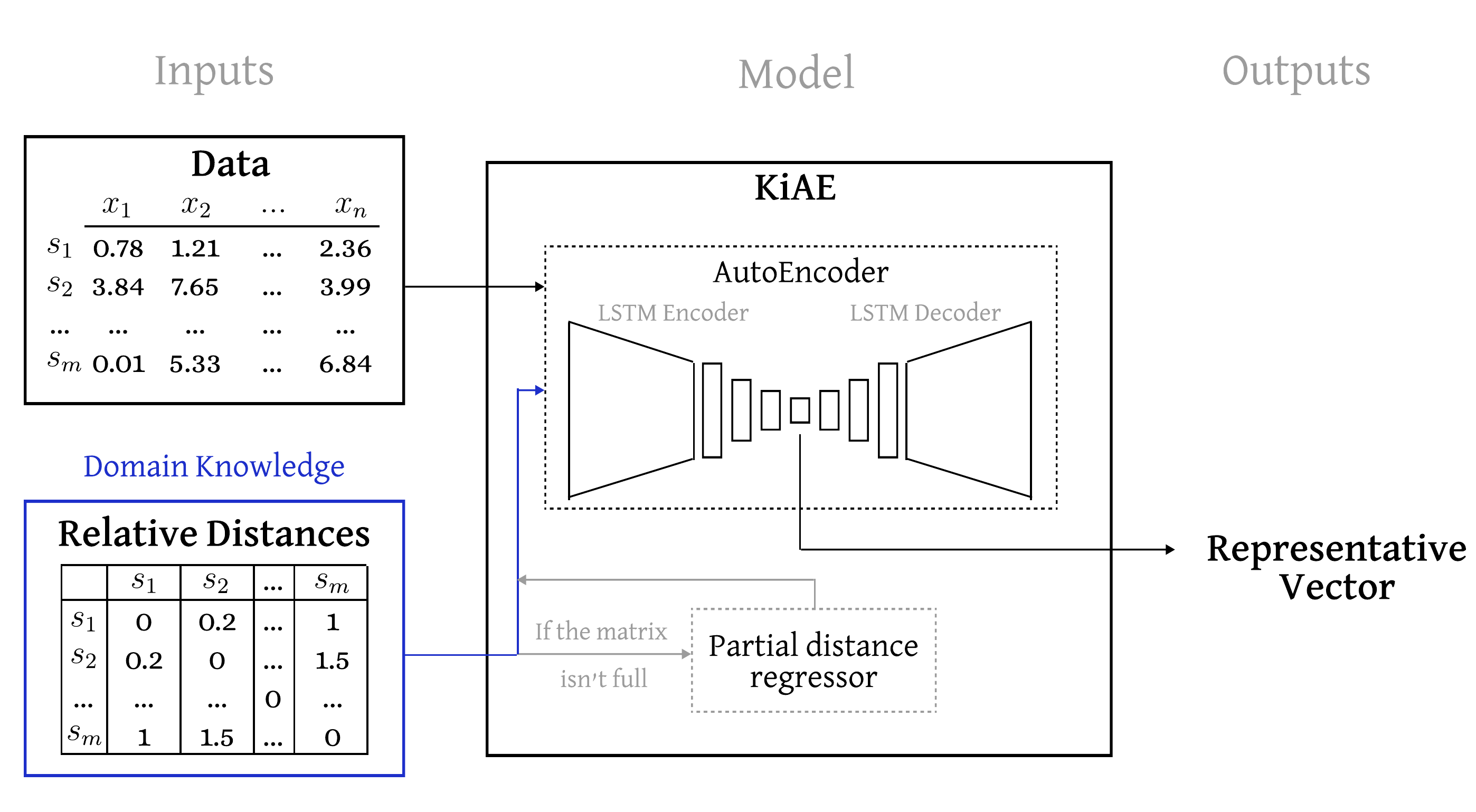}
    \caption{A schematic view of the proposed method, divided into its components and the interactions between them.}
    \label{fig:methods}
\end{figure}

\methodName{} is constructed from two components: a partial distance regressor and an LSTM-based AE with seven fully connected, size adaptive, layers (three as part of the encoder, three as part of the decoder, and one as a representation layer). Fig.~\ref{fig:methods} shows a schematic view of \methodName{}, presenting the inputs with domain-specific knowledge, the two components of \methodName{}, and the resulting outcome of a representative vector for each sample.

Using the two components, the proposed method operates in a \textit{training} phase and an \textit{inference} phase. During the training phase, the model obtains a tabular dataset of samples and a matrix of known distances between pairs of features (\(M_T\)). The matrix \(M_T\) is user-defined, where each value \(m_{i,j} \in [0,\infty)\) of the matrix represents an assumption of the relative distance between sample \(i\) and sample \(j\), compared to the distance between other pairs of samples. If the user does not have sufficient theoretical knowledge to fill \(M_T\) completely, then the matrix is passed into the partial distance regressor component (\(DR\), where the missing distance information is filled. Formally, \(DR\) is trained on the available distances of the pairs with a set of metrics provided by the user. Once \(DR\) is obtained, the empty entries are filled using an inference of the corresponding samples. Alternatively, the user can decide to leave the empty entries in \(M_T\) which will be ignored later on. Either way, once \(M_T\) is obtained, the AE model is trained to learn a lower-dimensional representation of the dataset. The loss function used for its training is a joint evaluation of the reconstruction capabilities of the model, and the capability to preserve all the distance properties portrayed in \(M_T\) in the \textit{Latent space}. Thus, the loss function is defined as follows:
\begin{equation}
    L(\{m_1, m_2, \dots, m_n\}) := \frac{1}{n^2 - n} \Big ( \omega_1 \sum_{i=1}^n \sum_{i=j}^n \big ( ||m_i - \Bar{m_i}|| \big ) + \omega_2 \sum_{i=1}^n  \big ( | ||R(m_i) - R(m_j)|| -  M_T(i, j)| \big )  \Big ),  
    \label{eq:loss_function}
\end{equation}
where \(m_i\) is a the \(i_{th}\) sample, \(\Bar{m_i}\) is the reconstructed \(i_{th}\) sample, \(R(m_i)\) is the representing vector of \(i_{th}\) sample, and \(\omega_1, \omega_2 \in [0, 1]\) such that \(\omega_1 + \omega_2 = 1\) are the weights of the domain-knowledge loss compared to the classical reconstruction loss of the AE. 

During the inference phase, as in other AE models, the decoder part of the AE is removed and the trained encoder is used to encode inputted data.

The AE used in \methodName{} is constructed of an encoder network that uses a bidirectional LSTM to produce an initial embedding, either capturing temporal (ordered) data or not. The decoder network uses a unidirectional LSTM to reconstruct the input sequence using the latent embedding only. The fully connected (FC) layers are used to further reduce the representation layer size and learn high-order non-linear connections in the data. The choice to use three FC layers was motivated by the fact that a larger number of FC layers is able to capture more complex dynamics on the one hand, but it requires more time, data, and computational resources to train efficiently. All FC layers are followed by a ReLU activation function. The model is trained using the Adam optimizer \cite{adam}, batch size of 16 samples, and 10 epochs. 

The AE's architecture can be easily altered to obtain better results for each specific task and dataset. Specifically, if the inputted sample's dimension, \(|m_i|\), is larger than a pre-defined length \(|m_i| > L \in \mathbb{N}\) than a sliding window of size \(L\) and a jump \(\omega \in \mathbb{N}^+\) is used. At the reconstruction end, a majority vote is performed to obtain the final prediction for the reconstructed sample.

\subsection{Experimental setup}

To determine the contribution of knowledge integration into an AE using the \methodName{} model, we carried out three experiments on datasets from three distinctly different scientific fields: economics, physics, and biology. As seen in Table.~\ref{tbl:dataset_attributes}, each of these fields is characterized by fundamentally different characteristics of feature space and common samples count.

\begin{table}[!ht]
\centering
\begin{tabular}{l l c c c}
\hline
 \textbf{Measure} &  & \textbf{Economics} & \textbf{Physics} & \textbf{Biology} \\ 
 \hline
 \textbf{Num. of features} & Commonly & \(O(10^0-10^1)\) & \(O(10^1-10^2)\) & \(O(10^{4}-10^{8})\) \\
                           & Our data & 9 & 33 & \(\sim 4.5 \cdot 10^6\) \\
 \hline
 \textbf{Num. of samples} & Commonly & \(O(10^4-10^9)\) & \(O(10^2-10^3)\) & \(O(10^1-10^2)\)\\
                          & Our data & 49627 & 2500 & 90 \\
 \hline
 \textbf{Num. of clusters} & Our data & 4 & 2 & 3 \\ \hline
 \textbf{Encoding space dimension} & Our data & 2 & 4 & 8192 \\
\end{tabular}
\caption{Characteristic feature space and sample count for datasets of three scientific fields, alongside the specific measures of the data used in this work. In addition, the number of known clusters and encoding space dimension in our data for each dataset.}
\label{tbl:dataset_attributes}
\end{table}

Specifically, to test out \methodName{} on data from these scientific fields, we used three example datasets that are available online. The economic dataset~\footnote{\url{https://www.carrefour.com/en}} contains pricing and other properties describing 49,673 products sold in a large supermarket. The products are divided into four main categories: premium, semi-premium, regular, and under-priced \cite{economy_data_categories_1,economy_data_categories_2}. The physics dataset contains different scales and ratios used to describe the mechanism of spherical particles settling in the air while experiencing aerodynamic drag \cite{scimed}. The dataset contains samples of 2,500 different settling spheres, that are categorized by their density into two groups: light and heavy particles.
Finally, for the biological dataset, we used 90 whole genome sequences equally divided between three species: Homosapien (humans), Rhesus macaque (Macaca mulatta), and Pan troglodytes (Chimpanzee), taken from NIH\footnote{\url{https://www.ncbi.nlm.nih.gov/}} and PGP-UK\footnote{\url{https://www.personalgenomes.org.uk/}}. The sequence indicators are provided as supplementary material. 

To define \(M_T\) for each dataset, we integrated the following knowledge:
\begin{enumerate}
    \item The distance between a sample to itself is \(\alpha_{i_i}=0\), by definition.
    \item Samples of the same category are closer to one another than to samples from a different category. Therefore, the distance between two samples of the same category is randomly sampled from a uniform distribution between \(\alpha_1\) and \(\alpha_2\). We arbitrarily chose \(\alpha_1 = 0\) and \(\alpha_2 = 1\). 
    \item The distance between sample \(i\) and sample \(j\), which are from two different groups \(x\) and \(y\), respectively, are randomly sampled from a uniform distribution between \(\gamma_{xy}\) and \(\gamma_{xy}+1\), such that \(\forall x, y: \gamma_{xy} > \alpha_2\). The order of \(\gamma_{xy}\) is determined per dataset as it reflects the domain-specific knowledge of relative distances between the categories of the data.
    \item In the economic and physics datasets the order of \(\gamma_{ij}\) between each pair of categories is set to \(\gamma_{ij}=1\). In the biology dataset, as Chimpanzees (group 1) are closest to Macaca mulattas (group 2), we set \(\gamma_{12}=1\). As humans (group 3) are more similar to Chimpanzees than to Macaca mulattas, we set \(\gamma_{13}=2\) and \(\gamma_{23}=3\).
    \item For the Noisy \methodName{}, the matrix \(M_T\) is filled with values ranging between 0 and 1 at random with a uniform distribution. 
\end{enumerate}

Of note, with this approach we were able to use classification data previously tagged by experts during the creation of these datasets, to approximate a domain expert's knowledge without any actual knowledge of these domains. In a more realistic scenario, the distances in \(M_T\) will be defined by more precise domain knowledge.

Each dataset was used to train and evaluate five models: AE with domain knowledge (\methodName{}), AE without domain knowledge (AE), \methodName{} with faulty domain knowledge (Noisy \methodName{}), AE architecture obtained using the automatic machine learning library AutoKeras \cite{autokeras} (Auto-AE), and an AE commonly used in the specific type of problem (COMN-AE). Formally, for the COMN-AE, we trained seven common AE architectures from Ref.~\cite{auto_ae_soruce}, and reported the best outcome. The Auto-AE is obtained by using AutoKeras as a \textit{black-box} limited to 100 model training. This is meant to balance between the computational burden and the need to allow a thorough enough search process so that the Auto-AE can converge to a well-performing solution. For the \methodName{} and Noisy \methodName{} cases, we set \(\omega_1 = \omega_2 = 0.5\). 

After obtaining the model for each of the experiments, the representation vector is computed for every sample in the dataset. Afterward, we computed the clustering using the Ward hierarchical clustering algorithm \cite{clustering_method}. Using these clusters, we computed the misclassification metric \cite{misclassification_metric}, calculated as the number of incorrect predictions divided by the total number of predictions. Since there are \(K!\) options to map for mapping a set of clusters to \(k\) groups (i.e., the possible ways to order \(k\) groups in a row), we tested all of them and reported the best score from all the options.

\section{Results}
\label{sec:results}

In this section, we present the results of our experiments with different AE techniques, including the \methodName{} method proposed in this paper. Table.~\ref{table:results} presents the misclassification rate score achieved by each model. Intuitively, the results show that introducing faulty distance and relation assumptions results in a poor ability to cluster the data, as seen by the high misclassification rates over all datasets and test cases. In all test cases, the \methodName{} model resulted in the lowest misclassification rate, leading by a rate of 4\%-19\% over the rest of the models. Apart from the economic dataset, the Auto-AE outperforms the COMN-AE, which outperforms the simple AE used in \methodName{}. However, in the economic dataset, the ranking is slightly revered during the fit and train test, with the COMN-AE outperforming the Auto-AE by a rate of 2\%-3\%.

\begin{table}[!ht]
\centering
\begin{tabular}{llccccc}
\hline
\textbf{Dataset} & \textbf{Test (data\%)} & \textbf{AE} & \textbf{KiAE} & \textbf{Noisy KiAE} & \textbf{Auto-AE} & \textbf{COMN-AE} \\ \hline
\multirow{3}{*}{Biology} & Fit (100\%) & 0.36 & 0.00 & 0.72 & 0.19 & 0.26 \\
 & Train (80\%) & 0.31 & 0.00 & 0.68 & 0.15 & 0.27 \\
 & Test (20\%) & 0.33 & 0.11 & 0.74 & 0.26 & 0.33 \\ \hline
\multirow{3}{*}{Physics} & Fit (100\%) 
& 0.24 & 0.04 & 0.69 & 0.16 & 0.21 \\
 & Train (80\%) 
& 0.26 & 0.03 & 0.68 & 0.17 & 0.22 \\
 & Test (20\%)
& 0.32 & 0.05 & 0.72 & 0.23 & 0.27 \\ \hline
\multirow{3}{*}{Economy} & Fit (100\%) & 0.22 & 0.05 & 0.88 & 0.18 & 0.15 \\
 & Train (80\%) & 0.19 & 0.05 & 0.86 & 0.16 & 0.14 \\
 & Test (20\%) & 0.22 & 0.07 & 0.91 & 0.21 & 0.25 \\ \hline
\end{tabular}
\caption{The results of the experiment in terms of clustering preservation in the embedding space, divided into three dataset types and five AE configurations. The results achieved during training are presented as the mean of k-fold cross-validation with \(k = 5\).}
\label{table:results}
\end{table}
 
\begin{figure}[!ht]
    \centering
    \includegraphics[width=0.99\textwidth]{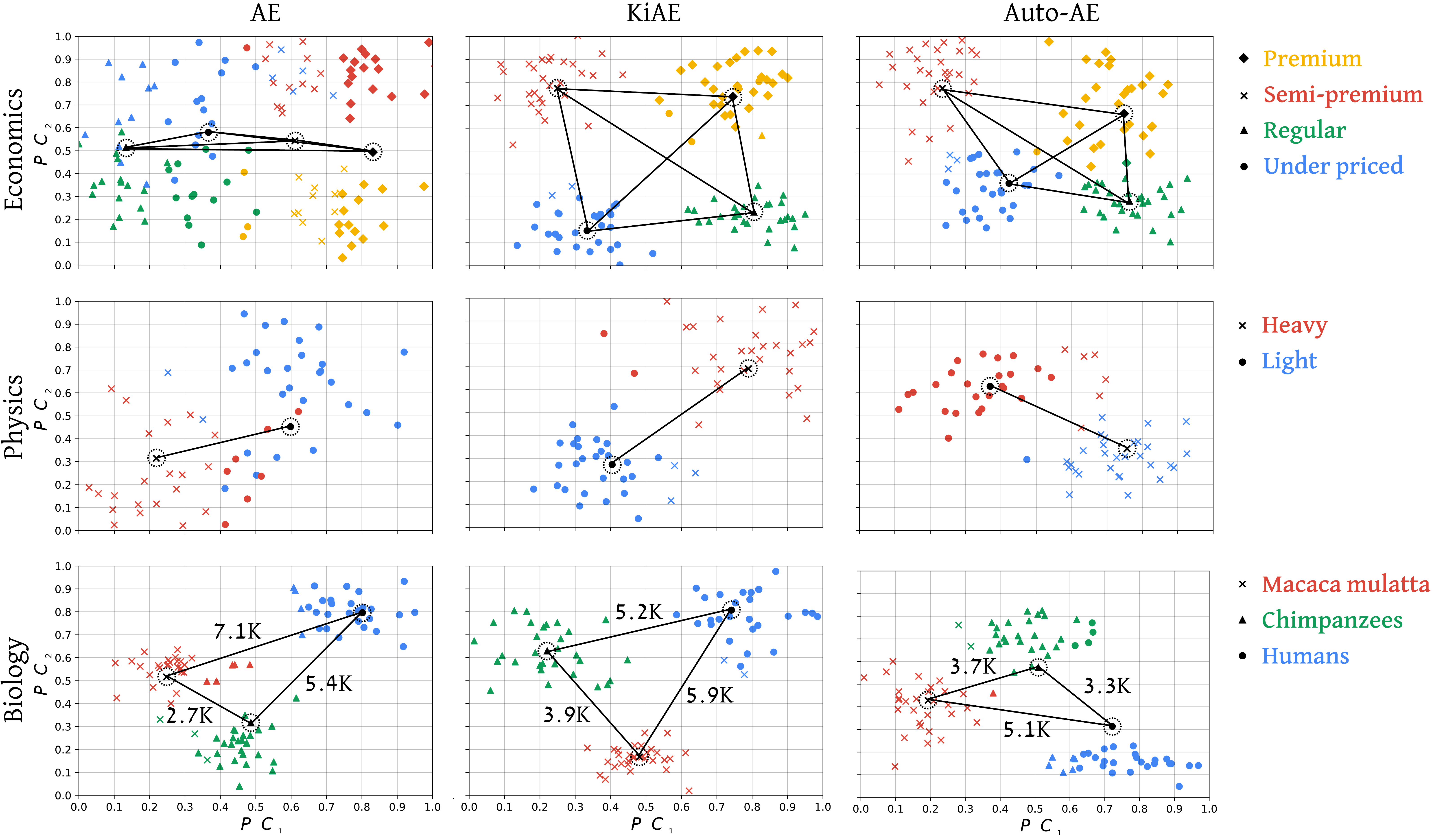}
    \caption{Comparison of the clusters detected by the AE, \methodName{}, and Auto-AE models for each dataset, for the testing cohort. The colors indicate the detected category while the marker shapes indicate the true label of the sample (e.g., in the physical dataset, blue circles are of samples correctly clustered as "light", while red circles were falsely clustered as "heavy"). The centers of each cluster are marked and connected. In the biology dataset (bottom row), the distances between each cluster center are reported as Euclidean distance in the \textit{latent space} (i.e. before performing PCA).}
    \label{fig:main_fig}
\end{figure}

Following these results, we visually explore how well the AE, \methodName{}, and Auto-AE were able to cluster the data. For that, we present Fig.~\ref{fig:main_fig}, which compares the performance of each technique (columns) on each dataset (rows). As the number of samples in the physics and economics datasets is too large to plot efficiently, we solved the set coverage task \cite{set_cover} with the greedy algorithm \cite{set_cover_algorithm} to find \(n = 90\) points with the smallest Euclidian distance to the obtained coverage points on the plot. This way, we ensure all the samples presented optimally represent the density and topology of the entire dataset, while keeping the visualized data small and consistent between the experiments \cite{set_cover_visual}. Afterward, we computed the PCA \cite{pca} of each dataset, presenting the two eigenvectors \(PC_1\) and \(PC_2\) with the largest in absolute value eigenvalues. 

Fig.~\ref{fig:main_fig} shows that the theoretical knowledge inserted to \methodName{} (middle column) is preserved in the \textit{latent space}. First, one can see that in the economic dataset (top), the centers of each cluster are set at roughly the same distance from one another, as was inserted. Second, in the physical dataset (middle), the centers of the two clusters are set the furthest apart, meaning that \methodName{} was able to create a better distinction between the two categories of the data. Third, in the biological dataset (bottom), both the AE and \methodName{} were able to preserve the theoretical distance ratios, while the Auto-AE failed (chimpanzees were found closest to humans rather than macaca mulattas).

\section{Discussion}
\label{sec:discussion}

In this research, we proposed a novel approach for integrating domain knowledge into an AutoEncoder (AE) neural network architecture to control the distance and neighborhood properties of the obtained latent space, in order to increase the accuracy of the models. Our method, called \methodName{}, incorporates two novel ideas into the AE architecture: (a) a meta-regression task that takes a partially fulfilled domain-knowledge matrix and fulfills it by training a regression model from the available entries; (b) a loss function that preserves neighborhood and distance based on the domain-knowledge.

The results presented in Table~\ref{table:results} show that \methodName{} outperforms a wide range of more advanced AE architectures by achieving significantly lower misclassification rates over three distinctly different scientific fields. The AE architectures examined and compared to \methodName{} included the automatic search of optimal AE architectures suitable to each dataset.

Additionally, in Fig.~\ref{fig:main_fig} we demonstrate that \methodName{} is able to capture both the distance and neighborhood properties of the samples in the \textit{latent space}. To demonstrate that the better performance of \methodName{} is associated with the new loss function rather than the AE architecture itself, we trained the architecture used for \methodName{}, which consistently showed worse results compared to \methodName{}. 

however, the results also show that the proposed method has some limitations. First, as in most knowledge-integrated models, the \methodName{}'s performance is highly dependent on the correctness and amount of domain knowledge provided to it. A poor quality (or even wrong) partial distance matrix can cause the model to obtain a mathematical result that complies with the inserted distance and neighborhood properties, yet harms the accuracy of the resulting clusters. To demonstrate this realistic scenario of inserting false relation assumptions, we compared in Table~\ref{table:results} the results obtained by \methodName{} with faulty domain knowledge. We show that this inserts a false bias into the model \cite{latest_1,latest_2,latest_3}, which results in the worst results of all the models tested. Second, finding the right dimension of the representation is a challenging task in this work done via a manual trial-and-error process by the authors. The naive approach to tackle this challenge is to perform a grid search on this parameter, which can become computationally expensive even for a small set of configurations \cite{grid_search}. Future work can try to find the optimal representation dimension given a dataset and previous experience using the meta-learning approach \cite{meta_1,meta_2,meta_3}. Third, further improvements can be made by investigating how \methodName{} can be extended to incorporate unlabeled data.

\bibliographystyle{unsrt}
\bibliography{bilbo}

\begin{thebibliography}{10}

\bibitem{regression_1}
X.~Jin, R.~Nie, D.~Zhou, S.~Yao, Y.~Chen, J.~Yu, and Q.~Wang.
\newblock A novel dna sequence similarity calculation based on simplified
  pulse-coupled neural network and huffman coding.
\newblock {\em Physica A: Statistical Mechanics and its Applications},
  461:325--338, 2016.

\bibitem{regression_2}
S.~Wang, F.~Tian, Y.~Qiu, and X.~Liu.
\newblock Bilateral similarity function: A novel and universal method for
  similarity analysis of biological sequences.
\newblock {\em Journal of Theoretical Biology}, 265(2):194--201, 2010.

\bibitem{regression_3}
S.~V. Buldyrev, N.~V. Dokholyan, A.~L. Goldberger, S.~Havlin, C-K. Peng, H.~E.
  Stanley, and G.~M. Viswanathan.
\newblock Analysis of dna sequences using methods of statistical physics.
\newblock {\em Physica A: Statistical Mechanics and its Applications},
  249(1-4):430--438, 1998.

\bibitem{regression_4}
X.~Yang and T.~Wang.
\newblock Linear regression model of short k-word: a similarity distance
  suitable for biological sequences with various lengths.
\newblock {\em Journal of Theoretical Biology}, 337:61--70, 2013.

\bibitem{regression_5}
B.~Liao, Q.~Xiang, L.~Cai, and Z.~Cao.
\newblock A new graphical coding of dna sequence and its similarity
  calculation.
\newblock {\em Physica A: Statistical Mechanics and its Applications},
  392:4663--4667, 2013.

\bibitem{regression_6}
B.~Liao, Q.~Xiang, L.~Cai, and Z.~Cao.
\newblock Numerical encoding of dna sequences by chaos game representation with
  application in similarity comparison.
\newblock {\em Genomics}, 108:134--142, 2016.

\bibitem{intro_phy_1}
M.~Raissi and G.~E. Karniadakis.
\newblock Hidden physics models: Machine learning of nonlinear partial
  differential equations.
\newblock {\em Journal of Computational Physics}, 357:125--141, 2018.

\bibitem{intro_phy_2}
R.~P. Springuel, M.~C. Wittmann, and J.~R. Thompson.
\newblock Reconsidering the encoding of data in physics education research.
\newblock {\em Phys. Rev. Phys. Educ. Res.}, 15:020103, 2019.

\bibitem{intro_eco_1}
X.~Su, Y.~Liu, and C.~Choi.
\newblock A blockchain-based p2p transaction method and sensitive data encoding
  for e-commerce transactions.
\newblock {\em IEEE Consumer Electronics Magazine}, 9(4):56--66, 2020.

\bibitem{intro_eco_2}
C.~Jiang, J.~Li, W.~Wang, and W-S. Ku.
\newblock Modeling real estate dynamics using temporal encoding.
\newblock In {\em Proceedings of the 29th International Conference on Advances
  in Geographic Information Systems}, page 516–525, 2021.

\bibitem{data_encoding}
N.~Yu, Z.~Li, and Z.~Yu.
\newblock Survey on encoding schemes for genomic data representation and
  feature learning—from signal processing to machine learning.
\newblock {\em Big Data Mining and Analytics}, 1(3):191--210, 2018.

\bibitem{ls_1}
C-K. Yeh, W-C. Wu, W-J. Ko, and Y-C.~F. Wang.
\newblock Learning deep latent space for multi-label classification.
\newblock {\em Proceedings of the AAAI Conference on Artificial Intelligence},
  31(1), 2017.

\bibitem{ls_2}
C.~Gelada, S.~Kumar, J.~Buckman, O.~Nachum, and M.~G. Bellemare.
\newblock {D}eep{MDP}: Learning continuous latent space models for
  representation learning.
\newblock In {\em Proceedings of the 36th International Conference on Machine
  Learning}, volume~97, pages 2170--2179, 2019.

\bibitem{ls_3}
A.~Scheinker.
\newblock Adaptive machine learning for time-varying systems: low dimensional
  latent space tuning.
\newblock {\em Journal of Instrumentation}, 16(10):P10008.

\bibitem{ga_dna_1}
R.~J. Parsons, S.~Forrest, and C.~Burks.
\newblock Genetic algorithms, operators, and dna fragment assembly.
\newblock {\em Machine Learning}, 21:11--33, 1995.

\bibitem{similarity_general_2}
A.~J. Pinho, D.~Parats, and S.~P. Garcia.
\newblock {GReEn}: a tool for efficient compression of genome resequencing
  data.
\newblock {\em Nucleic Acids Research}, 40(4):e27, 2012.

\bibitem{speech}
O.~Abdel-Hamid, A-R. Mohamed, H.~Jiang, L.~Deng, G.~Penn, and D.~Yu.
\newblock Convolutional neural networks for speech recognition.
\newblock {\em IEEE/ACM Transactions on Audio, Speech, and Language
  Processing}, 22(10):1533--1545, 2014.

\bibitem{nlp}
D.~Dligach, M.~Afshar, and T.~Miller.
\newblock {Toward a clinical text encoder: pretraining for clinical natural
  language processing with applications to substance misuse}.
\newblock {\em Journal of the American Medical Informatics Association},
  26(11):1272--1278, 2019.

\bibitem{latent_1}
P.~D. Hoff, A.~E. Raftery, and M.~S. Handcock.
\newblock Latent space approaches to social network analysis.
\newblock {\em Journal of the American Statistical Association},
  97(460):1090--1098, 2002.

\bibitem{latent_2}
F.~Monay and D.~Gatica-Perez.
\newblock On image auto-annotation with latent space models.
\newblock In {\em Proceedings of the Eleventh ACM International Conference on
  Multimedia}, page 275–278. Association for Computing Machinery, 2003.

\bibitem{latent_3}
S.~Rongali, A.~J. Rose, D.~D. McManus, A.~S. Bajracharya, A.~Kapoor,
  E.~Granillo, and H.~Yu.
\newblock Learning latent space representations to predict patient outcomes:
  Model development and validation.
\newblock {\em J Med Internet Res}, 22(3):e16374, 2020.

\bibitem{latent_define}
A.~Voynov and A.~Babenko.
\newblock Unsupervised discovery of interpretable directions in the {GAN}
  latent space.
\newblock In {\em Proceedings of the 37th International Conference on Machine
  Learning}, volume 119, pages 9786--9796. PMLR, 2020.

\bibitem{scimed}
Liron~Simon Keren, Alex Liberzon, and Teddy Lazebnik.
\newblock A computational framework for physics-informed symbolic regression
  with straightforward integration of domain knowledge.
\newblock {\em Scientific Reports}, 13(1):1249, 2023.

\bibitem{sr_1}
Maya Gupta, Andrew Cotter, Jan Pfeifer, Konstantin Voevodski, Kevin Canini,
  Alexander Mangylov, Wojciech Moczydlowski, and Alexander Van~Esbroeck.
\newblock Monotonic calibrated interpolated look-up tables.
\newblock {\em The Journal of Machine Learning Research}, 17(1):3790--3836,
  2016.

\bibitem{sr_2}
Gabriel Kronberger, Fabricio~Olivetti de~Fran{\c{c}}a, Bogdan Burlacu,
  Christian Haider, and Michael Kommenda.
\newblock Shape-constrained symbolic regression—improving extrapolation with
  prior knowledge.
\newblock {\em Evolutionary Computation}, 30(1):75--98, 2022.

\bibitem{auto_sklearn}
M.~Feurer, A.~Klevin, K.~Eggensperger, J.~T. Springenberg, M.~Blum, and
  F.~Hutter.
\newblock {\em Auto-sklearn: Efficient and Robust Automated Machine Learning}.
\newblock 2019.

\bibitem{ml_k_1}
R.~Wu, Y.~Fujita, and K.~Soga.
\newblock Integrating domain knowledge with deep learning models: An
  interpretable ai system for automatic work progress identification of natm
  tunnels.
\newblock {\em Tunnelling and Underground Space Technology}, 105:103558, 2020.

\bibitem{ml_k_2}
X.~Pan and H.~B. Shen.
\newblock Rna-protein binding motifs mining with a new hybrid deep learning
  based cross-domain knowledge integration approach.
\newblock {\em BMC Bioinformatics}, 18:136, 2017.

\bibitem{ml_k_3}
A.~Behnaz, H.~Kevin, and C.~Haoxian.
\newblock Interpretable feedback for automl and a proposal for
  domain-customized automl for networking.
\newblock In {\em Proceedings of the Twentieth ACM Workshop on Hot Topics in
  Networks}, page 53–60. Association for Computing Machinery, 2021.

\bibitem{manual_encoding}
L.~A. Chylek, L.~A. Harris, C-S. Tung, J.~R. Faeder, C.~F. Lopez, and W.~S.
  Hlavacek.
\newblock Rule-based modeling: a computational approach for studying
  biomolecular site dynamics in cell signaling systems.
\newblock {\em WIREs Systems Biology and Medicine}, 6(1):13--36, 2014.

\bibitem{geo_info}
E.~Winarno, W.~Hadikurniawati, and R.~N. Rosso.
\newblock Location based service for presence system using haversine method.
\newblock In {\em 2017 International Conference on Innovative and Creative
  Information Technology (ICITech)}, pages 1--4, 2017.

\bibitem{driven_encoding}
H.~Jiang, C.~Liu, J.~Paparrizos, A.~A. Chien, J.~Ma, and A.~J. Elmore.
\newblock Good to the last bit: Data-driven encoding with codecdb.
\newblock In {\em Proceedings of the 2021 International Conference on
  Management of Data}, page 843–856, 2021.

\bibitem{rule_data_encoding}
X-H. Li, C.~C. Cao, Y.~Shi, W.~Bai, H.~Gao, L.~Qiu, C.~Wang, Y.~Gao, S.~Zhang,
  X.~Xue, and L.~Chen.
\newblock A survey of data-driven and knowledge-aware explainable ai.
\newblock {\em IEEE Transactions on Knowledge and Data Engineering},
  34(1):29--49, 2022.

\bibitem{en_g_1}
M.~AlQuraishi and P.K. Sorger.
\newblock Differentiable biology: using deep learning for biophysics-based and
  data-driven modeling of molecular mechanisms.
\newblock {\em Nature Methods}, 18:1169--1180, 2021.

\bibitem{en_g_2}
L.~Himanen, A.~Geurts, A.~S. Foster, and P.~Rinke.
\newblock Data-driven materials science: Status, challenges, and perspectives.
\newblock {\em Advanced Science}, 6(21):1900808, 2019.

\bibitem{ae_review}
G.~Dong, G.~Liao, H.~Liu, and G.~Kuang.
\newblock A review of the autoencoder and its variants: A comparative
  perspective from target recognition in synthetic-aperture radar images.
\newblock {\em IEEE Geoscience and Remote Sensing Magazine}, 6(3):44--68, 2018.

\bibitem{rw_economy_1}
X.~Cui, S.~Liu, Z.~Lin, J.~Ma, F.~Wen, Y.~Ding, L.~Yang, W.~Guo, and X.~Feng.
\newblock Two-step electricity theft detection strategy considering economic
  return based on convolutional autoencoder and improved regression algorithm.
\newblock {\em IEEE Transactions on Power Systems}, 37(3):2346--2359, 2022.

\bibitem{rw_economy_2}
M.~Adkisson, J.~C. Kimmell, M.~Gupta, and M.~Abdelsalam.
\newblock Autoencoder-based anomaly detection in smart farming ecosystem.
\newblock In {\em 2021 IEEE International Conference on Big Data (Big Data)},
  pages 3390--3399, 2021.

\bibitem{rw_phy_1}
X.~He, Q.~He, and J-S. Chen.
\newblock Deep autoencoders for physics-constrained data-driven nonlinear
  materials modeling.
\newblock {\em Computer Methods in Applied Mechanics and Engineering},
  385:114034, 2021.

\bibitem{rw_phy_2}
T.~Finke, M.~Krämer, A.~Morandini, and I.~Oleksiyuk.
\newblock Autoencoders for unsupervised anomaly detection in high energy
  physics.
\newblock {\em Journal of High Energy Physics}, page 161, 2021.

\bibitem{rw_phy_3}
M.~Farina, Y.~Nakai, and D.~Shih.
\newblock Searching for new physics with deep autoencoders.
\newblock {\em Physical Review D}, 101:075021, 2020.

\bibitem{rw_bio_1}
J.~Marino.
\newblock {Predictive Coding, Variational Autoencoders, and Biological
  Connections}.
\newblock {\em Neural Computation}, 34(1):1--44, 2022.

\bibitem{rw_bio_2}
D.~Chicco, P.~Sadowski, and P.~Baldi.
\newblock Deep autoencoder neural networks for gene ontology annotation
  predictions.
\newblock In {\em Proceedings of the 5th ACM Conference on Bioinformatics,
  Computational Biology, and Health Informatics}, page 533–540. Association
  for Computing Machinery, 2014.

\bibitem{ae_info_intro}
Y.~Deng, A.~Sander, L.~Faulstich, and K.~Denecke.
\newblock Towards automatic encoding of medical procedures using convolutional
  neural networks and autoencoders.
\newblock {\em Artificial Intelligence in Medicine}, 93:29--42, 2019.

\bibitem{ae_info_1}
T.~Ma and A.~Zhang.
\newblock Integrate multi-omics data with biological interaction networks using
  multi-view factorization autoencoder (mae).
\newblock {\em BMC Genomics}, 20:944, 2019.

\bibitem{ae_info_2}
W.~Ding, H.~Lin, B.~Li, K.~J. Eun, and D.~Zhao.
\newblock Semantically adversarial driving scenario generation with explicit
  knowledge integration.
\newblock {\em arXiv}, 2022.

\bibitem{adam}
D.~P. Kingma and J.~Ba.
\newblock {Adam: A Method for Stochastic Optimization}.
\newblock {\em arXiv}, 2014.

\bibitem{economy_data_categories_1}
J.~Anselmsson, U.~Johansson, and N.~Persson.
\newblock Understanding price premium for grocery products: a conceptual model
  of customer‐based brand equity.
\newblock {\em Journal of Product \& Brand Management}, 16(6):401--414, 2007.

\bibitem{economy_data_categories_2}
R.~Volpe.
\newblock Evaluating the performance of u.s. supermarkets: Pricing strategies,
  competition from hypermarkets, and private labels.
\newblock {\em Journal of Agricultural and Resource Economics}, 36(3):488--503,
  2011.

\bibitem{autokeras}
H.~Jin, F.~Chollet, Q.~Song, and X.~Hu.
\newblock Autokeras: An automl library for deep learning.
\newblock {\em Journal of Machine Learning Research}, 24(6):1--6, 2023.

\bibitem{auto_ae_soruce}
F.~Chollet.
\newblock Building autoencoders in keras, 2016.
\newblock Accessed on March 8th, 2023.

\bibitem{clustering_method}
J.~H. Ward.
\newblock Hierarchical grouping to optimize an objective function.
\newblock {\em Journal of the American Statistical Association},
  58(301):236--244, 1963.

\bibitem{misclassification_metric}
J.~E. Chacon.
\newblock A close-up comparison of the misclassification error distance and the
  adjusted rand index for external clustering evaluation, 2021.

\bibitem{set_cover}
E.~Balas and M.~W. Padberg.
\newblock On the set-covering problem.
\newblock {\em Operations Research}, 20(6):1152--1161, 1972.

\bibitem{set_cover_algorithm}
V.~Chvatal.
\newblock A greedy heuristic for the set-covering problem.
\newblock {\em Mathematics of Operations Research}, 4(3):233--235, 1979.

\bibitem{set_cover_visual}
R.~Hu, T.~Sha, O.~Van~Kaick, O.~Deussen, and H.~Huang.
\newblock Data sampling in multi-view and multi-class scatterplots via set
  cover optimization.
\newblock {\em IEEE Transactions on Visualization and Computer Graphics},
  26(1):739--748, 2020.

\bibitem{pca}
K.~F. R.~S. Pearson.
\newblock Liii. on lines and planes of closest fit to systems of points in
  space.
\newblock {\em The London, Edinburgh, and Dublin Philosophical Magazine and
  Journal of Science}, 2(11):559--572, 1901.

\bibitem{latest_1}
O.~L. Liu, H-S. Lee, C.~Hofstetter, and M.~C. Linn.
\newblock Assessing knowledge integration in science: Construct, measures, and
  evidence.
\newblock {\em Educational Assessment}, 13(1):33--55, 2008.

\bibitem{latest_2}
A.~Best, J.~L. Terpstra, G.~Moor, B.~Riley, C.~D. Norman, and R.~E. Glasgow.
\newblock Building knowledge integration systems for evidence‐informed
  decisions.
\newblock {\em Journal of Health Organization and Management}, 23(6):627--641,
  2009.

\bibitem{latest_3}
A.~Tiwana.
\newblock An empirical study of the effect of knowledge integration on software
  development performance.
\newblock {\em Information and Software Technology}, 46(13):899--906, 2004.

\bibitem{grid_search}
R.~Liu, E.~Liu, J.~Yang, M.~Li, and F.~Wang.
\newblock Optimizing the hyper-parameters for svm by combining evolution
  strategies with a grid search.
\newblock {\em Intelligent Control and Automation. Lecture Notes in Control and
  Information Sciences}, 344, 2006.

\bibitem{meta_1}
M.~Huisman, J.~N. van Rijn, and A~Plaat.
\newblock A survey of deep meta-learning.
\newblock {\em Artificial Intelligence Review}, 54:4483–4541, 2021.

\bibitem{meta_2}
R.~Vilalta, C.~Giraud-Carrier, and P.~Brazdil.
\newblock {\em Meta-Learning - Concepts and Techniques}, pages 717--731.
\newblock Springer US, 2010.

\bibitem{meta_3}
R.~Vilalta and Y.~Drissi.
\newblock A perspective view and survey of meta-learning.
\newblock {\em Artificial Intelligence Review}, 18:75--95, 2002.

\end{thebibliography}

\section*{Declarations}
\subsection*{Funding}
This research did not receive any specific grant from funding agencies in the public, commercial, or not-for-profit sectors.

\subsection*{Conflicts of interest/Competing interests}
The authors have no financial or proprietary interests in any material discussed in this article.

\subsection*{Materials availability}
The data that has been used is presented in the manuscript with the relevant sources and available with the code on the project's page on GitHub.

\subsection*{Author Contributions}
Teddy Lazebnik: Conceptualization, formal analysis, investigation, methodology, software, supervision, and writing - original draft. \\ 
Liron Keren-Simon: Conceptualization, formal analysis, visualization, writing - original draft, and writing - review \& editing.

\subsection*{Acknowledgements}
The authors wish to thank Walter Lutz, Alexander Liberzon, and Labib Shami for helping with the data gathering for the experiments and Stephan Beck and Ismail Moghul for helping with the research conceptualization.

\end{document}